\documentclass{article}

\usepackage{microtype}
\usepackage{graphicx}
\usepackage{booktabs} 

\usepackage{hyperref}

\usepackage{multirow}
\usepackage{url}
\usepackage[colorinlistoftodos]{todonotes}
\usepackage{subcaption}
\usepackage{hhline}
\usepackage{comment}
\usepackage{slashbox}
\usepackage{mathtools}

\newcommand{\hs}[1]{{\color{blue}{#1}}}

\newcommand{\Harsh}[1]{{\color{blue} Harsh: #1}}
\newcommand{\edits}[1]{{#1}}

\usepackage[accepted]{icml2019}

\icmltitlerunning{AntMan: Sparse Low-Rank Compression to Accelerate RNN inference}

\begin{document}

\twocolumn[
\icmltitle{AntMan: Sparse Low-Rank Compression to Accelerate RNN inference}


\icmlsetsymbol{equal}{*}

\begin{icmlauthorlist}
\icmlauthor{Samyam Rajbhandari}{equal,to}
\icmlauthor{Harsh Shrivastava}{equal,goo}
\icmlauthor{Yuxiong He}{to}
\end{icmlauthorlist}

\icmlaffiliation{to}{Microsoft Research, Redmond, USA.}
\icmlaffiliation{goo}{Computer Science \& Engineering department, Georgia Insititure of Technology, USA.}

\icmlcorrespondingauthor{Samyam Rajbhandari}{samyamr@microsoft.com}
\icmlcorrespondingauthor{Harsh Shrivastava}{hshrivastava3@gatech.edu}

\icmlkeywords{Machine Learning, ICML}

\vskip 0.3in
]



\printAffiliationsAndNotice{\icmlEqualContribution} 

\begin{abstract}
Wide adoption of complex RNN based models is hindered by their inference performance, cost and memory requirements. To address this issue, we develop AntMan, combining structured sparsity with low-rank decomposition synergistically, to reduce model computation, size and execution time of RNNs while attaining desired accuracy.
AntMan extends knowledge distillation based training to learn the compressed models efficiently.
Our evaluation shows that AntMan offers up to 100x computation reduction with less than 1pt accuracy drop for language and machine reading comprehension models. Our evaluation also shows that for a given accuracy target, AntMan produces 
5x smaller models than the state-of-art. 
Lastly, we show that AntMan offers super-linear speed gains compared to theoretical speedup, demonstrating its practical value on commodity hardware. 
\end{abstract}
\section{Introduction} \label{sec:intro}
Remarkable advances in deep learning (DL) have produced great models across a wide variety of tasks such as computer vision, machine reading, speech generation and image recognition \citep{goodfellow2016deep}. However, wide adoption of these models is still limited by their inference performance, cost and memory requirements. On the client side, all pervasive devices like smart-phones, tablets and laptops have limited memory and computational resources to handle large complex DL models. On the server side, intensive computation can render the models too slow to meet responsiveness requirements and too expensive to scale, preventing their deployment in production.

Model Compression is a flourishing area that aims to reduce the computational and memory complexity of DL models to address the aforementioned problems without significantly affecting accuracy.  Compressing Convolution Neural Networks (CNNs) have already been widely explored in the past few years \citep{cheng2017survey}, while our work focuses on Recurrent Neural Networks (RNNs), which are broadly used among various natural language processing tasks  \citep{mikolov2010recurrent,seo2016bidirectional,zaremba2014recurrent}. It is well known that large RNN models are  computation and memory intensive \citep{zhang2018deepcpu}. In particular, their computation increases linearly with sequence length, and their recurrent unit has to be computed sequentially, one step at a time with limited parallelism, both of which makes long execution time a crucial issue for RNN inference computation.
Compressing RNNs, however, is challenging, because a recurrent unit is shared across all the time steps in sequence, compressing the unit will aggressively affect all the steps.

Inducing sparsity is one of the prominent approaches used for RNN compression. \citet{narang2017exploring} proposed a pruning approach that deletes up to 90\% connections in RNNs.  The obtained sparse matrices, however, have an irregular/non-structured pattern of non-zero weights, which is unfriendly for efficient computation in modern hardware systems \citep{lebedev2016fast,wen2016learning}. To address this issue, \citet{narang2017block} proposed inducing block-sparsity in RNNs via pruning or group lasso regularization. Similarly, \citet{wen2017learning} introduces ISS, intrinsic structured sparsity for LSTMs \citep{hochreiter1997long}, a type of RNN , such that a sparse LSTM can be transformed into a dense one but with smaller size.  ISS conveniently turns sparsity into efficient execution, but as its sparse structure is quite coarse-grained, it is hard to push out high sparsity without degrading accuracy, especially in RNNs where the hidden dimension is smaller than input dimension (elaborated in Section~\ref{sec:comp-reduction}).    

Our work explores a new line  of structured sparsity on RNNs, using predefined compact structures as opposed to pruning and regularization based approaches.
We take inspiration from predefined compact CNN structures such as group convolutions \citep{ZhangZLS17,krizhevsky2012imagenet} and depth-wise separable convolutions \citep{chollet2017xception}. Specifically, we replace matrix-vector multiplications (MVs), the dominant part of RNN computations, with localized group projections (LGP). LGP divides the input and output vectors into groups where the elements of the output group are computed as a linear combination of those from the corresponding input group. 
In addition, to empower the information flow across multiple groups along the steps of RNN computation, we use a permutation matrix or a dense-square matrix to combine outputs across groups, helping the compact structure to retain accuracy.

Additionally, we combine LGP with low-rank matrix decomposition to further reduce the computations. 
This is possible as low rank and sparsity are complimentary to each other. Low-rank decomposition such as SVD approximates a low-rank multiplication $A \textbf{x}$ as $P Q \textbf{x}$, where $P$ and $Q$ are dense. By imposing LGP-based sparsity on $P$ and $Q$, we reduce the computation further. For a given rank reduction factor of $r$, we reduce the computation cost and model size by $\mathcal{O}(r^2)$, compared to $\mathcal{O}(r)$ by using low-rank decomposition methods like SVD \citep{golub1970singular} alone.   

We call our compression approach AntMan --- `shrink in scale' by synergistically combining structured sparsity and low-rank decomposition, but `increase in strength' by enabling the flow across structured groups along RNN sequence to retain accuracy. 

To train RNN models with AntMan, we use teacher-student training paradigm \citep{bucilua2006model} by combining the label loss with teacher-MSE-loss and teacher-KL-divergence-loss. 
To improve the training efficiency, we develop a new technique to decide proper coefficients to obtain high accuracy efficiently with minimal trials.

We evaluate AntMan on multiple RNN based models for machine reading comprehension and language modeling. For a well-known MRC model \citep{seo2016bidirectional}, we reduce the computational complexity and model size of LSTMs (a particular type of RNN) by up to 25x with less than 1pt drop in F1 score. For PTB \citep{zaremba2014recurrent} language model, we achieve a computational reduction of 50x with no drop in perplexity, and 100x with just a single point drop in perplexity. We also construct language models for PTB with perplexities ranging from 64 to 70, but with 3x to 5x fewer overall model weights (5x to 25x reduction in RNN weights) than the state-of-art.

Lastly, we develop efficient implementations of inference kernels on CPUs to serve models compressed by AntMan. We show that unlike computation with unstructured sparsity, AntMan offers significant performance improvement for large RNN models even with modest levels of sparsity. Our evaluations show that a 2x to 10x theoretical reduction in computation can result in up to 2x to 30x actual speedup, respectively, for moderate to large RNNs, demonstrating attractive practical value of AntMan on commodity hardware.
\section{Related Work}
\label{sec:related-work}
\textbf{Compressing RNNs via Sparsity:} Discussed in Sec.~\ref{sec:intro}.

\textbf{Compressing RNNs via Low-Rank Approximations and Parameter Sharing:} 
\citet{prabhavalkar2016compression,lu2016learning} use SVD  to compress LSTM models by 3-4x for acoustic modeling and speech recognition tasks with negligible loss in accuracy. 
\edits{ \cite{lu2016learning} also demonstrate parameter sharing to reduce the memory complexity of RNNs, without reducing the computational cost.}  
AntMan compresses both memory and computation, and it achieves significantly higher compression rate than the low-rank based methods without losing accuracy.
\cite{Ye2017LearningCR} uses Block Tensor Decomposition (BTD) to compress LSTMs for vision tasks. Their work is specifically designed to exploit redundancies present in the image vector (input to the LSTMs) obtained from upstream CNN layers, while AntMan is designed to compress general RNNs, where the inputs do not come from CNNs and do not exhibit such redundancies in many cases. \edits{Furthermore, BTD is designed to compress only the input vector and not the hidden vectors. This hinders the performance of BTD over a range of RNNs, where the hidden vectors are also large. }

\edits{
\textbf{Quantization} \cite{deep-compression} reduces the model complexity by using lower precision for both computation and weights. \cite{lee2018retraining} show that for RNN based language models, 2-3 bit quantization can be used to compress the model with minimal perplexity loss. However, it should be noted that modern CPUs only support up to 8-bit quantization \cite{lower-precision-inference} which limits the compression to at most 4x. Furthermore, quantization is complimentary to AntMan. It can be applied to AntMan, or any full-precision compressed model to further reduce the compute and memory complexity. }

\textbf{Teacher-Student training paradigm:} Knowledge Distillation (KD) developed by \citet{hinton2015distilling} is a popular technique to compress deep and wide networks into sparser ones, where the compressed model mimics the distribution learned by the complex model. \edits{\citet{oguntola2018slimnets} show that compression techniques such as pruning, and low-rank decomposition can be combined with KD to significantly improve compression rate, while maintaining accuracy.}

KD usually optimizes a weighted average of \emph{two} different objective functions. The first objective function can be one of the following three: cross entropy, or mean square error, or Kullback Leibler divergence, all computed with respect to the soft targets, and the second objective function is the cross entropy with the correct labels. Several similar approaches \citep{romero2014fitnets,luo2016face,chen2015net2net,balan2015bayesian,zagoruyko2016paying} extend the idea of KD. 

In contrast, AntMan combines \emph{three} objective functions, MSE loss, KL divergence loss and the cross entropy of the true labels, powered by an efficient method of deciding their coefficients. 
\begin{table*}[t!]
    \begin{minipage}[c]{\textwidth}
    \centering
    \resizebox{0.85\textwidth}{!}{
        \centerline{\includegraphics[width=\columnwidth]{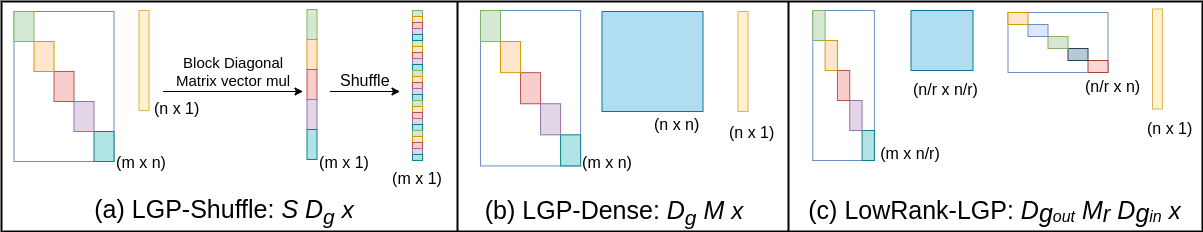}}
        }
        \caption{Three different AntMan modules to compress $A x$.}
        \label{bdsh}
        \vskip -0.30in
    \end{minipage}
\vspace{-2mm}
\end{table*}
\section{AntMan Design and Implementation} \label{sec:am-compression}
AntMan compresses RNN computation by combining benefits of structured sparsity and low rank decomposition. It consists of three components: i) localized group projections that sparsify matrix multiplications using block diagonal matrices, ii) group mixing that exchanges information across different local groups along the sequence of RNN computation, and iii) low rank approximation that uses SVD like decomposition to reduce the rank of the projection.  By composing them, we construct a few variations of AntMan compression modules that exhibit varying degree of compression rate and accuracy impact.
We also analyze the cost complexity of AntMan modules and discuss efficient implementation on commodity hardware such as CPUs using off-the-shelf BLAS libraries.

\subsection{Localized Group Projections}
\label{sec:shuffle}

AntMan reduces the computation and size of RNNs by replacing dense matrix-vector product (MV) with sparse but structured MV.
It divides the input and output vectors into $g$ local groups such that the elements of an output group is a weighted linear sum of the elements in the corresponding input group.  Since output elements of one group only depend on the input elements of the corresponding group, we call it localized group projections (LGP). Mathematically, we replace the matrix vector product $A \textbf{x}$ with $D_g \textbf{x}$, where $D_g$ is a block-diagonal matrix with $g$ blocks.

In a RNN cell computation, the hidden-state vector at time-step $t-1$ is an input to the MV used to compute the hidden-state vector at time-step $t$. Therefore,   
 using LGP to replace MV in RNN restricts the information flow within a single local group across multiple time steps of the RNN. 
This restriction reduces the expressibility of the RNN, potentially degrading accuracy. AntMan uses `group mixing' to address this issue.

\subsection{Group Mixing}
To facilitate the information flow across multiple   localized groups along RNN sequence computation, AntMan multiplies the output (or input) vector of LGP with a square matrix, which we call mixing matrix.  
We develop two types of mixing with varying memory and computational complexity --- shuffle mix and dense mix --- inspired by the shuffling layer \citep{ZhangZLS17} used with group convolutions, or 1x1 convolutions used with depth-separable convolutions \citep{chollet2017xception}. 

\textbf{Shuffle mix:} The shuffle-mix matrix is a permutation matrix, which
evenly distributes the elements of the same group across the entire output vector across different groups.  
Figure~\ref{bdsh}a shows the effect of shuffle mix following LGP.
Mathematically, shuffle mix is equivalent to a transpose operation. If the output vector $\textbf{v}$ resulting from the block diagonal MV has $m$ elements, we can represent the vector as a matrix $O$ of size $[g, m/g]$, where each row represents an output group computed from the corresponding input group. Shuffle mix simply transforms $\textbf{v}$ to $O^T$ of shape $[m/g, g]$.

\textbf{Dense mix:} This technique uses a dense square matrix for group mixing
when the matrix in the MV is non-square. 
Mathematically, given $A \textbf{x}$, where size of $A$ is $m$ x $n$, we can decompose it into $M D_g \textbf{x}$, when $m <n$, or $D_g M\textbf{x}$, when $n < m$,  and $M$ is a dense-mixing matrix of size $m$ x $m$, or $n$ x $n$, respectively. Figure~\ref{bdsh}b shows an example of dense mix preceding LGP. 

Dense mix has added cost of the dense matrix vector multiply compared to shuffle mix (quadratic vs linear). However, unlike shuffle mix that simply permutes the elements of the output vector, dense mix takes a weighted linear combination, making it more general.  It helps retain accuracy at the expense of additional computation. When combined with low-rank decomposition discussed next, dense mix provides high compression while maintaining accuracy, which we elaborate further in evaluation (Table~\ref{tab:cost-complexity}).


\subsection{Low-rank Decomposition}
Low-rank decomposition such as SVD approximates a low-rank matrix-vector product $A \textbf{x}$ as $P Q \textbf{x}$, where $A$, $P$ and $Q$ are dense with shapes $m$ x $n$, $m$ x $\frac{n}{r}$ and $\frac{n}{r}$ x $n$, respectively, and $\frac{n}{r}$ is the reduced rank. We combine it with LGP by adding LGP-based sparsity on $P$ and $Q$, further reducing computation.
This combination is likely to obtain more compression than using either of the techniques alone because structured sparsity and low-rank decomposition operate in a complimentary fashion. In particular, low rank reduces the computation by factorizing $A$ into smaller matrices $P$ and $Q$, while LGP reduces computation by sparsifying these matrices without changing their dimensions. 

\subsection{Assembling AntMan Compression Modules}
Composed from the three components, \emph{LGP}, \emph{group mixing} and \emph{low-rank decomposition}, we construct variations of AntMan compression modules to address varying efficiency and accuracy demand across DL models.  Figure~\ref{bdsh} shows three of such compression modules: (a) LGP-shuffle --- LGP with shuffle mix; (b) LGP-dense --- LGP with dense mix; (c) LowRank-LGP --- low rank with LGP-dense.

We elaborate the compression modules by taking Figure~\ref{bdsh}(c), LowRank-LGP, as an example. LowRank-LGP combines structured sparsity with low rank decomposition.  First, it   
decomposes an MV into an SVD-like form, i.e., $A \textbf{x} \leftarrow P Q \textbf{x}$, where $A$ is a matrix of size $m$ x $n$, $P$ and $Q$ are decomposed matrices of size $m$ x $\frac{n}{r}$ and $\frac{n}{r}$ x $n$, and $\frac{n}{r}$ represents the reduced rank. Next, we  replace $P$ and $Q$ using LGP-Dense, i.e., $A\textbf{x} \leftarrow D_{g_{out}} M_{out} M_{in} D_{g_{in}} \textbf{x} $ , where $D_{g_{in}}$ and $D_{g_{out}}$ are block-diagonal matrices of size $m$ x $\frac{n}{r}$ and $\frac{n}{r}$ x $n$, and $g_{in}$ and $g_{out}$ are the number of diagonal blocks. $M_{in}$ and $M_{out}$ are both square matrices of size $\frac{n}{r}$ x $\frac{n}{r}$. It can be further simplified into $A\textbf{x} \leftarrow D_{g_{out}} M_r D_{g_{in}} \textbf{x}$, where $M_r = M_{out} M_{in}$. This module, combining all three components of AntMan, exhibits the potential of achieving significantly higher cost reduction than using SVD alone, which we quantify shortly.

\subsection{Computation and Model Size Reduction}

\begin{table}
    \centering
    \resizebox{0.48\textwidth}{!}{
    \begin{tabular}{c|c|c|c}
    \textbf{Name} & {\bf Computation} & {\bf Model Size} &{\bf Cost Reduction}\\
    \hline
    \parbox[][1cm][c]{0.9cm}{ Matrix-Vector} & $A \textbf{x}$ & $mn$ & 1 \\
    \hline
    \parbox[][1cm][c]{0.9cm}{ SVD} & $P_r Q_r \textbf{x}$ & $\dfrac{mn}{r} + \dfrac{nn}{r}$ &  $\dfrac{mr}{m + n}$ \\
    \hline
    \parbox[][1cm][c]{0.9cm}{ LGP-Shuffle} & $S D_g \textbf{x}$ & $\dfrac{mn}{g}$ & g \\
    \hline
    \parbox[]{0.9cm}{LGP Dense}& \parbox[][1.5cm][c]{1.9cm}{\centering $M D_g \textbf{x}$ or\\ $D_g M \textbf{x}$}  &\parbox[][][c]{1.9cm}{\centering $\dfrac{mn}{g} + min(m,n)^2$ } & \parbox[][][c]{2.9cm}{\centering $\dfrac{mn}{\dfrac{mn}{g} + min(m,n)^2}$} \\
    \hline
    \parbox[][2cm][c]{1.2cm}{ LowRank -LGP}& $D_{g_{out}} M_r D_{g_{in}} \textbf{x}$ & \parbox[][2cm][c]{1.9cm}{\centering $\dfrac{mn}{rg_{out}} + \dfrac{nn}{rg_{in}} + \dfrac{n^2}{r^2}$ } & $\dfrac{mr^2g_{out}g_{in}}{mg_{in}r + ng_{out}r + ng_{out}g_{in}}$ \\
    \end{tabular}
    }
    \caption{Comparison of model computation and size reduction.}
    \label{tab:cost-complexity1}
\vspace{-4mm}
\end{table}

Table~\ref{tab:cost-complexity1} discusses the reduction in computation and model size over the original $A \textbf{x}$, where $A$ is a matrix of size $m$ x $n$.
The third column reports the total number of multiply-add operations, which is also the size of weight matrix in the case of MV.  
The final column represents the reduction in computation (that is equal to the reduction in model size) compared to the original MV.

We highlight two key messages: (1)
LGP-Dense reduces the total cost by $\approx \frac{max(m,n)}{min(m,n)}$ when $g\gg \frac{max(m,n)}{min(m,n)}$, i.e., the larger difference between $m$ and $n$, the more reduction it gets. 
(2) 
When $g_{out}$ and $g_{in}$ are large enough, LowRank-LGP can enable significantly higher cost reduction over SVD, while maintaining the same reduced rank.
To see this, let's assume $m=n$, and $g_{out} = g_{in} = g$. In this case, the cost reduction from SVD is $r/2$, while the cost reduction from LowRank-LGP is $\frac{r^2g}{2r + g}$. Now, if $g \geq r$, then the cost reduction is at least $r^2/3$, and it goes up to $r^2$ when $g \gg r$. Therefore, the reduction in computational cost scales as $O(r)$ for SVD, while it scales as $O(r^2)$ for LowRank-LGP assuming $g \geq r$. 

As a concrete example, consider a MV of size $1000$x$400$, where the number of parameters and the number of multiply-add (MADD) operations in 400K. Using LGP-Shuffle with $g=10$, we can reduce both the number of parameters and MADD operations to 40K. Using LGP-Dense with $g=10$, we can reduce them to 200K (40K from LGP + 160K from dense mix). Using LowRank-LGP with $g=10$ and $r=4$, we can reduce the parameters and MADD operations to $\frac{1000*400}{4*10} + \frac{400*400}{4*4} + \frac{400*400}{4 *10}$, which is 24K. 
 
\subsection{Efficient Implementation}
\label{sec:efficient-implementation}
We develop efficient implementation of AntMan modules (LGP-Shuffle, LGP-Dense, LowRank-LGP) on CPUs to empower their usage in practice. The implementation consists of three building blocks: i) Regular matrix-vector multiply for dense mix, ii) shuffle-mix multiply and iii) block-diagonal MVs. BLAS libraries such as Intel MKL already provides efficient implementation of matrix-vector multiplication. Shuffle mix is implemented efficiently as a matrix transpose operation as described in Section~\ref{sec:shuffle}. The block-diagonal MV is viewed as multiple smaller MVs, each corresponding to one of the blocks in the block-diagonal matrix. With multicores, each of these blocks can be computed in parallel using OpenMP for parallelization and Intel MKL for MV computation. In summary, AntMan modules can be implemented efficiently on commodity hardware, such as CPUs, conveniently applicable to various devices on cloud and on edge.

\section{Training AntMan Using Knowledge Distillation} \label{section:kt}
We observe that while training AntMan models directly on target labels alone does not generalize well on test data, using knowledge distillation or teacher-student training helps greatly on retaining accuracy. We use the original uncompressed model as the teacher, and train the compressed model (student) to imitate the output distribution of the teacher, in addition to training on the target labels.  We describe how we apply and extend teacher-student training.

\textbf{Loss function:} We define the loss function of the compressed model as a weighted combination of \emph{three} losses --- the raw loss from the target labels, and the MSE and the KL divergence losses of the student's output distribution with respect to the teacher's corresponding output distribution: 
\begin{equation}\label{total_loss}
\begin{split}
    \text{Loss}_{total} &= C_{target}\times \text{Loss}_{target}(S_o, T_{target})  +\\  & C_{mse} \times \text{Mse}(S_o, T_o) + C_{kl} \times \text{KL}(S_o, T_o)
\end{split}
\end{equation}
where $C_{target}, C_{mse}, C_{kl}$ are the coefficient values corresponding to the target loss, MSE loss and KL divergence loss, respectively. $S_o, T_o$ are the output distributions of student and teacher model, respectively, whereas $T_{target}$ is the target distribution. 
\textbf{Deciding loss coefficients:} The final performance of the compressed model significantly depends on the values of the loss coefficients, $C_{target}, C_{mse}$ and $C_{kl}$. Searching for appropriate values for these coefficients via grid or random search is time and resource consuming.  
We develop an efficient method to decide them with the following intuition. The direction of the gradient updates to the model is dictated by the relative magnitudes of the individual losses. If one is significantly smaller than the other, then it would have minimal impact on the direction of the updates during training. Therefore, we want to scale each of the three losses such that the overall magnitude of each of the three terms in Eqn.~\ref{total_loss} is roughly the same. To this end, we initially train the compressed model separately using each of the three losses and record the loss values once the model converges. Then we use these values as reference to identify loss coefficients such that each of the three terms is roughly the same. We use these coefficients to train the compressed model to optimize accuracy.

\textbf{Effectiveness:} We demonstrate the effectiveness of our approach by training a 10x compressed language model constructed by replacing the LSTMs in \citet{zaremba2014recurrent} with LGP-Shuffle with $g=10$. For this compressed model, the validation loss values at convergence when training separately with the three individual losses were: target = $4.110$, MSE = $0.133$ and KL = $0.004$. 

Table~\ref{grid_coeff} shows the test perplexity values (lower the better) obtained by training the compressed model with varying $C_{mse}$ and $C_{KL}$, while fixing $C_{target}=1$. Note that the lowest perplexity is achieved by setting coefficient values of $C_{target}=1$, $C_{mse}=30$ and $C_{KL}=1000$. At these values, each term in Eqn.~\ref{total_loss} is roughly equal to $4$, demonstrating the effectiveness of our method.

Table~\ref{grid_coeff} also shows the benefits of combining \emph{three} losses. Note that when $C_{mse} =0$, the best achievable perplexity is $121.97$. Similarly, when $C_{KL}=0$, the best achievable perplexity is $75.61$. However, combining all three gives the lowest perplexity of $74.69$. 

\begin{table}
\centering
\resizebox{0.45\textwidth}{!}{
      \begin{tabular}{|@{\hspace*{2mm}}c|*{5}{c}}\hline
      \backslashbox{$\bf C_{mse}$}{$\bf C_{kl}$}
      &\makebox[1em]{0}&\makebox[1em]{1}&\makebox[1em]{100}
      &\makebox[1em]{\bf 1000}&\makebox[1em]{10000}\\\hline
      $0$		&172.25	&135.87	&131.47	&121.97	&127.66 \\
      $1$		&91.65	&91.60	&90.98	&90.81	&91.67\\
      $\bf 30$	&75.61	&81.43	&75.47	&\bf{74.69}	&75.39\\
      $100$	&76.91	&84.00	&76.65	&76.72	&76.84\\
      $500$	&78.73	&86.29	&78.88	&78.63	&78.87\\
      \end{tabular}}
\caption{Different choices of coefficients vs test perplexities for student model with 10x computation reduction on the PTB dataset.}
\label{grid_coeff}
\vspace{-4.5mm}
\end{table}

\section{Experiments}
\label{sec:exp}
We evaluate AntMan on three aspects. 
(1) We use AntMan to obtain order(s) of magnitude computation reduction for language modeling and machine reading comprehension tasks while getting similar accuracy. 
(2) We use AntMan to construct models with several times fewer parameters than the state-of-art models with the same accuracy targets.
(3) Not limited by theoretical speedup, we measure real speedup of AntMan on CPUs and observe super-linear computational efficiency on large RNNs, demonstrating attractive practical value of AntMan on commodity hardware.

\begin{table*}[tb!]
    \begin{minipage}[c]{\textwidth}
    \footnotesize
\begin{center}
\resizebox{0.99\textwidth}{!}{
          \begin{tabular}{ccccccccccc}
          \multicolumn{1}{p{1.4cm}}{\bf Compute Reduction}
          &\multicolumn{1}{p{1.1cm}}{\bf weights\# LSTMs}
          &\multicolumn{1}{p{1.1cm}}{\bf \centering ISS}
          &\multicolumn{1}{p{1.1cm}}{\bf \centering Pruning + KD$_s$}
          &\multicolumn{1}{p{1.1cm}}{\bf \centering Pruning + KD$_o$}
          &\multicolumn{1}{p{1.1cm}}{\bf \centering LowRank + KD$_s$}
          &\multicolumn{1}{p{1.1cm}}{\bf \centering LowRank + KD$_o$}
          &\multicolumn{1}{p{1.4cm}}{\bf \centering Direct Design+KD$_s$}
          &\multicolumn{1}{p{1.4cm}}{\bf \centering Direct Design+KD$_o$}
          &\multicolumn{1}{p{1.1cm}}{\bf \centering LowRank-LGP}
          &\multicolumn{1}{p{1.1cm}}{\bf \centering LGP-Shuffle}
          \\ \hline \\
           10x	&3.60M &78.65 &80.02 &75.61 &114.79 &78.63 &107.84 &80.06 &75.99 &\textbf{74.69}\\ 
          50x	&0.72M &- &86.80 &78.86 &125.27 &81.40 &129.07 &99.96	&79.11	&\textbf{77.38}\\ 
          100x	&0.36M &- &91.81 &81.21 &148.78 &88.59 &155.10 &129.3 &81.27 &\textbf{78.66}\\ 
          \end{tabular}
          }
          \end{center}
          \caption{Computation reduction of models on PTB data and test perplexity values. Compression is applied to the large LSTM model \cite{zaremba2014recurrent} having a test perplexity of \textbf{77.55} with \textbf{36M} LSTM parameters.}
          \label{tab: ptb78}
    \end{minipage}
\vspace{-1mm}
\end{table*}

\subsection{Computation Reduction} \label{sec:comp-reduction}
We evaluate the effectiveness of AntMan on reducing model computation:
(1) on \citet{zaremba2014recurrent} PTB model for word level completion task, we obtain $50$x reduction without sacrificing any accuracy; (2) on \citet{seo2016bidirectional} BiDAF model for machine reading compression task, we obtain up to  $25$x reduction with less than 1pt drop on F1 score. See footnote\footnote{For both PTB and BiDAF, the optimizer and all the hyper-parameters for training the respective models were set as default from their standard and recommended implementations on github 
\cite{ptb-github, bidaf-github} 
Training was completed on single a NVIDIA Titan XP GPU in few hours for PTB and less than 2 days for BiDAF. AntMan was implemented in PyTorch (for PTB) and Tensorflow (for BiDAF) in just a few hundred lines of code. We will release the code once the paper is accepted.} for experimental details.  
\subsubsection{Word level completion}
\label{sec:word-level-completion}
Word level completion task predicts the next word given a partial input sequence. 

\textbf{Dataset and Model:}
We use Penn Tree Bank(PTB) dataset \citep{marcus1993building} that consists of 929k training words, 73k validation words and 82k test words. 
As the teacher model, we chose the model in \citet{zaremba2014recurrent} consisting of 2 layered LSTMs each with hidden dimension of 1500. \edits{For the student model, we replace all the MVs in the LSTMs with i) LGP-Shuffle ($10$x, $g$=$10$; $50$x, $g$=$50$; $100$x, $g$=$100$), and ii) LowRank-LGP ($10$x, $r$=$10$, $g$=$10$; $50$x, $r$=$20$, $g$=$25$; $100$x, $r$=$40$, $g$=$15$).}

\edits{
\textbf{Techniques in Comparison}: We compare AntMan with various compression techniques as shown in Table~\ref{tab: ptb78}: 

\emph{ISS} leverages intrinsic structured sparsity of LSTMs
\cite{wen2017learning}.
 We take the results directly from the paper to compare with.  ISS does not use knowledge distillation.  

\emph{Pruning+KD$_s$} combines pruning with knowledge distillation using KL-divergence as described in SlimNets \cite{oguntola2018slimnets}.  \emph{Pruning+KD$_o$} applies pruning together with our version of knowledge distillation (KD$_o$) that combines KL-divergence with MSE-loss.
In both cases, we prune $90/98/99\%$ of weights in the LSTMs to achieve a 10/50/100x reduction in the LSTM computation.

\emph{LowRank+KD$_s$} and \emph{LowRank+KD$_o$} 
apply a low-rank form ($Ax$=$PQx$) to the MVs in the LSTMs, 
and train the model using KD$_s$ \cite{oguntola2018slimnets} and KD$_o$ respectively. For each MV, we apply low-rank decomposition by choosing a rank ($r$) such that there is 10/50/100x reduction in computation. $r$ can be found using the formula $r = \frac{m \times n}{x \times{(m+n)}}$, where $m$ and $n$ are rows and columns of the weight matrix, and $x$ is the reduction factor. 

\emph{Direct Design+KD$_s$/KD$_o$} use a reduced hidden-dimension ($h'$) in the LSTMs such that the computation in the LSTMs is 10/50/100x less than the original LSTM. $h'$ can be found by solving the equation $(i+h')\times h' = \frac{(i+h)\times h}{x}$ where, $i$, $h$, $h'$ and $x$ are input dimension, old hidden dimension, new hidden dimension, and computation reduction factor, respectively.  
}

\begin{table}
\centering
        \begin{center}
        \bgroup
        \def\arraystretch{1.40}
    \scriptsize
        \begin{tabular}{|c|c|c|}
        \hline
        \multicolumn{1}{|c}{\multirow{2}{5em}{\bf Theoretical Compression}}  & \multicolumn{2}{|c|}{\bf Actual Performance Gain} \\
        \hhline{~--}
        & \multicolumn{1}{c|}{\bf Pruning} & \multicolumn{1}{c|}{\bf LowRank, ISS, AntMan} \\
        \hhline{---}
		10x	&2x	& 9-11x\\
        \hline
        \end{tabular}
        \egroup
        \end{center}
    \caption{Theoretical vs Actual Performance of compression techniques on large LSTM model \cite{zaremba2014recurrent} for PTB. Here, theoretical refers to the reduction in computation, while actual refers to the speedup observed during execution.}   
    
    \label{tab:baseline-prun}
\vspace{-3mm}
\end{table}

\textbf{Results:} Table~\ref{tab: ptb78} shows the perplexity values of the compressed models for different levels of computation reductions. 
While matching the original model's perplexity, AntMan with LGP-Shuffle ($g=50$) achieves $50$x computation reduction.
It achieves 10x computation reduction ($g=10$) with 3pt better perplexity and achieves $100$x computation reduction ($g=100$) with only 1pt perplexity loss. 
\edits{
Among all the techniques, AntMan (with LGP-Shuffle) achieves the best perplexity at all levels of compression.

Also note our knowledge distillation (KD$_o$) approach that combines KL-divergence with MSE loss does much better at retaining accuracy of the compressed models than the conventional knowledge distillation approach described in \citet{oguntola2018slimnets} (see KD$_s$ vs KD$_o$ in Table.~\ref{tab: ptb78}), further validating the effectiveness of our KD$_o$ approach.
}

\edits{
\textbf{Compute Efficiency in Practice:} Table~\ref{tab:baseline-prun} shows the actual performance gain in execution time \footnote{\edits{The pruned model was executed using Intel \texttt{mkl\_scsrmm} to run the sparse matrix-multiply (MM), and Intel \texttt{cblas\_gemm} was used to execute the dense MM in all the other compressed models.}} with the 10x compressed models trained using pruning, ISS, low-rank, and AntMan.  While ISS, low-rank and AntMan all achieve an actual performance gain of approximately 10x, pruning is limited at 2x due to reduced computation efficiency of unstructured sparsity (eg. difficulty in using vector processing on unstructured data). This demonstrates the limited practical value of unstructured pruning for model acceleration.

 Quantization can provide a decent memory compression (Table \ref{tab:quantization}), achieving 79.42 perplexity at 16x memory reduction with 2-bit quantization. However, AntMan does significantly better, achieving nearly 1pt lower perplexity even at 100x memory reduction (Table~\ref{tab:baseline-prun}). Furthermore, the compute efficiency of quantized model depends on hardware support. On modern CPUs, 8-bit and 16-bit compression are supported by the hardware architecture, resulting up to 4x performance gains, but anything below 8-bit are not well-supported, making it practically impossible to take advantage of (see Sec.~\ref{sec:related-work}). 

\begin{table}
\centering
        \begin{center}
        \bgroup
        \def\arraystretch{1.45}
    \scriptsize
        \begin{tabular}{|c|c|c|c|}
        \hline
        \multicolumn{1}{|c}{\multirow{2}{7em}{\bf (Full model)\\ Test Perplexity}}  & \multicolumn{3}{|c|}{\bf Model Size} \\
        \hhline{~---}
        & \multicolumn{1}{c|}{\bf 2-bit} & \multicolumn{1}{c|}{\bf 3-bit} & \multicolumn{1}{c|}{\bf 4-bit}  \\
        \hhline{----}
		78.27	&79.42	&76.94 &77.05\\
        \hline
        \end{tabular}
        \egroup
        \end{center}
    \caption{Quantization results on \citep{zaremba2014recurrent}'s large LSTM model. These results are taken from \cite{lee2018retraining}.} 
    \label{tab:quantization}
\vspace{-4mm}
\end{table}

}

\begin{table*}[t!]
    \begin{minipage}[c]{\textwidth}
    \footnotesize
    	\begin{center}
         \begin{tabular}{c|cc|cccccc|c}
\hline
\textbf{Description}&\textbf{EM} & \textbf{F1} & \textbf{ModFwd1} & \textbf{ModBwd1} & \textbf{ModFwd2} & \textbf{ModBwd2} & \textbf{OutFwd} & \textbf{OutBwd} & \textbf{weight\#}\\
\hline
Expert&67.9&77.3&1x&1x&1x&1x&1x&1x&2.69M\\
\hline
ISS&65.29&75.47&1.95x&2.26x&6.14x&4.34x&5.87x&8.85x&1.03M\\
\hline
\textbf{LowRank-LGP-noKD}&66.07&76.11&8.6x&8.6x&7.5x&7.5x&11.2x&11.2x&0.75M\\
\textbf{LGP-Shuffle}&65.41&75.87&9.09x&9.09x&6.66x&6.66x&9.09x&9.09x&0.78M\\
\textbf{LowRank-LGP 1}&66.06&76.73&12.5x&12.5x&9.09x&9.09x&16.66x&16.66x&0.69M\\
\textbf{LowRank-LGP 2}&65.86&76.6&20.42x&20.42x&17.7x&17.7x&25.39x&25.39x&0.56M\\
\hline
\end{tabular}
          \end{center}
          \caption{Comparison of computation reduction between AntMan and ISS for BiDAF}
          \label{tab:cost-complexity}
    \end{minipage}
\vspace{-2mm}
\end{table*}
\subsubsection{Machine Reading Comprehension (MRC)}
MRC tasks have gained significant popularity in last few years within NLP and computer vision communities. The models answer a query about a given context paragraph, evaluated based on exact match (EM) and F1 score (higher the better).

\textbf{Dataset: } We use Stanford Question Answering Dataset (SQuAD) \citep{rajpurkar2016squad}, which consists of a large set of Wikipedia articles and more than 100,000 questions. The answer to every question is always a small excerpt of the article.

\textbf{Teacher Model:} We chose our teacher model as the BiDirectional Attention Flow Model (BiDAF) \citep{seo2016bidirectional}, which is a hierarchical multi-stage model with 6 layers. We focus on compressing the layers having RNNs, which are also the most computationally expensive ones. 
Specifically, the modeling layer uses 2 layers of bi-directional LSTMs, denoted by ModFwd1, ModBwd1, ModFwd2, ModBwd2, while the output layer has a single bi-directional LSTM, denoted by OutFwd, OutBwd. 

\textbf{Compressed Models: } We created four compressed models  using AntMan with different levels of compression to replace the LSTMs in the BiDAF model: i) LGP-Shuffle $(g_{im}$=$10, g_{hm}$=$4)$, ii) LowRank-LGP 1 $(g_{im}$=$10, g_{hm}$=$5, r_{im}$=$4, r_{hm}$=$2)$, and iii) LowRank-LGP 2 $(g_{im}$=$5, g_{hm}$=$5, r_{im}$=$8, r_{hm}$=$4)$, and iv) LowRank-LGP-noKD $(g_{im}$=$5, g_{hm}$=$5, r_{im}$=$4, r_{hm}$=$2)$, trained without using knowledge distillation . Here, $g_{im}$ and $g_{hm}$ refers to the number of groups, and $r_{im}$ and $r_{hm}$ refers to the low-rank reduction factors for input and hidden MVs of the LSTMs, respectively. The computation reduction for each LSTM is shown in Table~\ref{tab:cost-complexity}.

\edits{
\textbf{Techniques in Comparison:} We use ISS as the primary baseline for this comparison as it is the best performing one among all baselines\footnote{ISS achieves better perplexity at 10x compression than all the baselines: Pruning+KD$_s$, LowRank+KD$_s$ and DirectDesign+KD$_s$. While Pruning+KD$_o$, does better using our knowledge distillation scheme, ISS achieves significantly better compute efficiency.} as shown in Sec.~\ref{sec:word-level-completion}. 
 
}

\textbf{Results:} Table~\ref{tab:cost-complexity} shows that 
AntMan (both LGP-Shuffle and LowRank-LGP) achieves significant computation reduction over the original.  Moreover, \edits{the reduction is much higher than ISS while achieving better EM and F1 scores.  
Furthermore, even without using knowledge distillation, we demonstrate the effectiveness of AntMan's compressed structure (LowRank-LGP-noKD) over ISS.   }   

To elaborate the results, ISS compresses an RNN by reducing hidden dimension. The amount of computation per LSTM step for ISS is proportional to $(i+h/r)*(h/r)$, where $i$ is the input dimension, $h$ is the hidden dimension, and $1/r$ is fraction of hidden dimension removed by ISS. When $i \gg h$, the compression is proportional to $r$. In BiDAF, $i\gg h$ in the first modeling layers (800 vs 100). Therefore, compression in these layers is proportional to the reduction in the hidden dimension. However, $h$=$100$ is already very small. By reducing it further, ISS experiences near 2pt drop in F1 score with less than 2.5x compression on the first modeling layers.

LGP-Shuffle uses structured sparsity to compress both the input and hidden MVs without reducing hidden dimension. For a comparable EM and F1 scores to ISS, LGP-shuffle achieves significantly higher reduction on the first modeling layers, while doing modestly better on all other layers. 
LowRank-LGP improves further upon LGP-Shuffle, increasing accuracy by leveraging dense mix to enrich the connection among multiple localized groups, and reducing computation by combining low-rank decomposition.  It achieves significantly higher computation reduction across all layers than both ISS and LGP-Shuffle, while achieving nearly 1pt higher F1 scores.
\subsection{Optimized model size for different accuracy targets}

Different applications have various accuracy requirements while the devices they are running on also impose different constraints on the model size. For given accuracy targets, smaller models are desired; and for given model sizes, higher accuracy is desired.  We show that AntMan improves the Pareto curve of model size against accuracy, providing more compressed models with the same accuracy targets of several recent models at word level completion task.  


\textbf{Teacher Model:} We use the state-of-art language model as of Dec, 2018, AWD-LSTM \citep{merity2017regularizing}, consisting of 3-layer LSTMs with 1150 hidden units and an embedding size of 400. 

\textbf{Compressed Models:} Our compressed models replace all the MVs in the LSTMs of AWD-LSTM with AntMan (LGP-Shuffle with $g$=$5$ to $g$=$50$ groups).

\begin{figure}[t]
\scriptsize
    \begin{center}
    \centerline{\includegraphics[width=0.99\columnwidth]{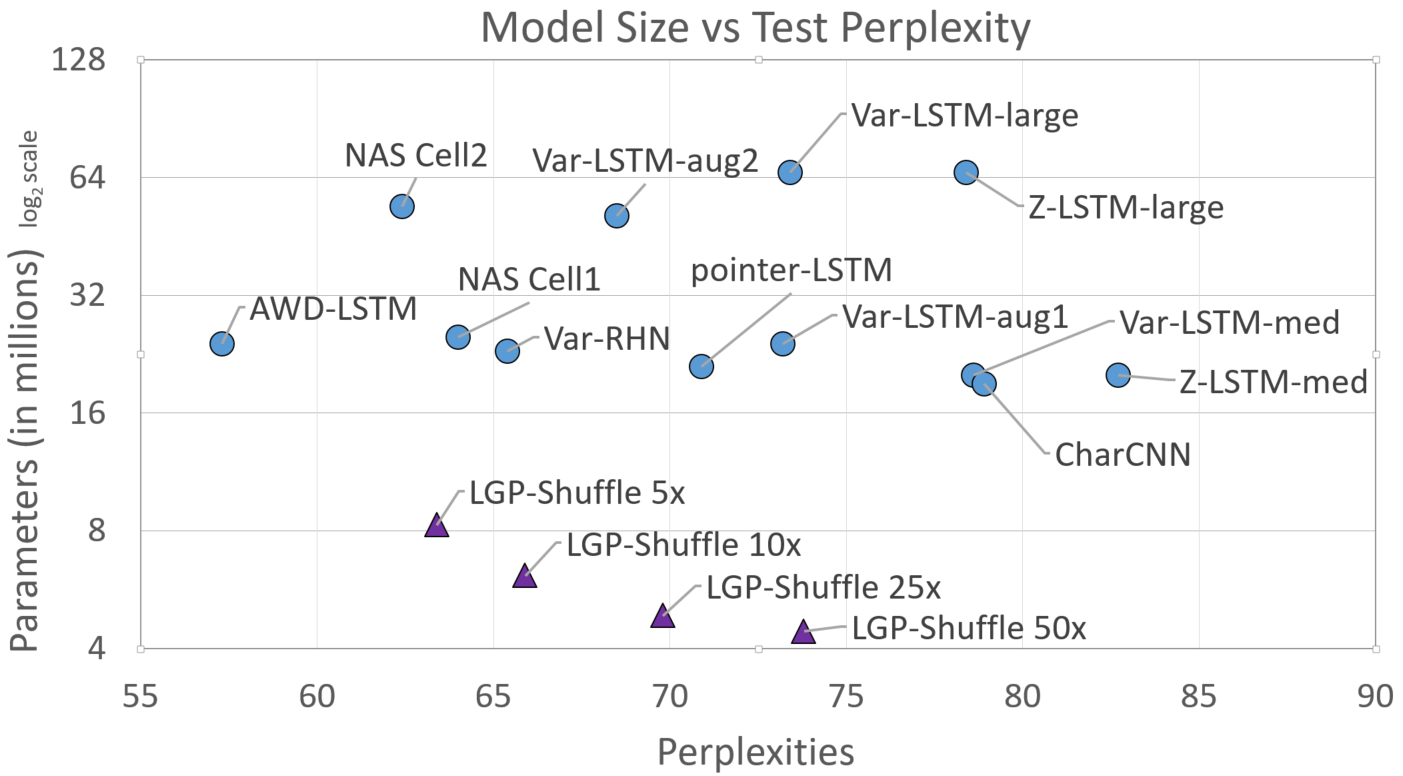}}
    \captionof{figure}{Comparing the number of model parameters vs perplexity of AntMan based models with various other language models published in the last four years, extracted from Table1 in \citep{merity2017regularizing}. The AntMan based models (LGP-Shuffle) are shown as purple triangles.}
    \label{fig:densePTB}
    \end{center}
\vspace{-8.5mm}
\end{figure}

\textbf{Results:} Figure~\ref{fig:densePTB} compares AntMan with other models. LGP-Shuffle ($g$=$5$) achieves perplexity of 63 with 5x fewer LSTM parameters and 3x fewer total parameters than NAS-Cell \citep{zoph2016neural}, the state-of-art model obtaining this range of accuracy. LGP-Shuffle ($g$=$10$) achieves perplexity of 66 with 10x fewer LSTM parameters and 4x fewer total parameters than Var-RHN \citep{zilly2016recurrent}, and LGP-Shuffle ($g$=$50$) achieves perplexity of 74 with 50x fewer LSTM parameters and 5x fewer total parameters than Var-LSTM-avg1 \citep{inan2016tying}. These results notably improve the Pareto curve of the task by reducing model sizes against different accuracy targets.\footnote{We did not aim to reproduce the state-of-art perplexity (57.3px at 24M parameters) of AWD-LSTM model. AWD-LSTM uses various regularization techniques, each with its own set of hyper-parameters, requiring extensive hyper-parameter tuning to reach its state-of-art perplexity. The AntMan results presented in Figure~\ref{fig:densePTB} was achieved without any regularization. Trying to match AWD-LSTM perplexity using AntMan with regularization could be an exercise in large scale hyper-parameter tuning, which is beyond the scope of this paper.}

\subsection{Theoretical vs Actual Speedup}\label{sec:compute-efficiency}
By efficiently implementing AntMan as described in
Section~\ref{sec:efficient-implementation}, we turn the theoretical speedup (computation reduction) to actual speedup (execution time reduction) in practice. Table~\ref{tab:actual-vs-theory} further shows that the actual speedup can be significantly higher than the theoretical speedup for large problem sizes. 

\textbf{Problem Configuration:}  We measure the execution time of LSTMs with and without AntMan by varying input and hidden dimensions from 100 to 1600. We use a batch size of 1, which is common in serving scenarios, and a sequence length of 100. 

\textbf{LSTM Implementation:} We use an efficient implementation as discussed in \citet{rnn-performance}: Fuse 4 input MVs across all time steps into a single large matrix multiplication, and fuse 4 hidden MVs within each time step. 

\textbf{Platform:} The experiments are run on a single core of Intel CPU E5-2650 v4 @ 2.20GHz. We use just a single core for two reasons: i) to emulate the limited resource availability in many use cases such as laptops and smart phones, ii) performance of multi-core RNN is highly implementation dependent \citep{zhang2018deepcpu} even for regular RNNs and therefore is difficult to make apple-to-apple comparison. We use Intel MKL library 
for GEMM implementation.   

\begin{table}
\centering
\resizebox{0.45\textwidth}{!}{
    \begin{tabular}{c|c|c|c|c|c}
        \hline
        \multicolumn{1}{c}{\multirow{3}{3.5em}{\bf Input \& Hidden Dim}} &\multicolumn{1}{|c}{\multirow{3}{3.2em}{\bf Memory (MB)}}  & \multicolumn{4}{|c}{\bf Actual vs Theoretical Speedup} \\
        \hhline{~~----}
        & & \multicolumn{2}{c|}{\bf LGP-Shuffle} & \multicolumn{2}{c}{\bf LowRank-LGP} \\
        \hhline{~~----}
        & & {\bf 2x} & {\bf 10x} & {\bf 2.66x} & {\bf 8x}\\
        \hhline{------}
        100&0.32&1.10x&1.16x&1.10x&0.08x\\
        400&5.12&1.89x&6.39x&2.32x&4.70x\\
        800&20.48&2.00x&8.66x&2.78x&6.50x\\
        1200&46.08&4.80x&24.02x&6.50x&20.00x\\
        1600&81.92&5.40x&30.20x&7.42x&23.80x\\
    \end{tabular}}
    \caption{Measured speedup on CPU using LGP-Shuffle and LowRank-LGP compared to the theoretical speedup for various input and hidden dimension. For LGP-Shuffle, we use $g$=$2$ and $g$=$10$ to get a theoretical speedup of 2x and 10x. For LowRank-LGP, we use $g$=$2$ and $r$=$2$, and $g$=$10$, and $r$=$2$ to get a speedup of 2.66x and 8x, respectively.} 
    \label{tab:actual-vs-theory}
\vspace{-4mm}
\end{table}

\textbf{Discussion:} Table~\ref{tab:actual-vs-theory} shows that, for very small problem size, AntMan offers no speedup regardless of the reduction in the computation. This is expected as GEMM performance gets worse as the problem size decreases \citep{zhang2018deepcpu,rajbhandari2017optimizing}. However, as the problem size is already very small, memory reduction or performance improvement is less crucial for such problems. For medium sized problems, AntMan offers good actual speedup compared to the theoretical speedup. Notice that unlike unstructured sparsity, where significant levels of sparsity is necessary to see actual performance improvement, with AntMan, even a modest 50\% sparsity or 2x computation reduction results in significant performance gain at problem size 400 and 800. Furthermore, for large problem sizes the actual speedup is significantly larger than the theoretical speedup. At problem size of 1200 and 1600, the weight matrices in the LSTM are too large to fit in L3 cache (30 MB in this case), thus spilling into memory. These LSTMs have much lower efficiency as  the memory bandwidth of a CPU is much lower than the L3 cache bandwidth. By reducing the memory footprint, AntMan-based LSTM fits in L3 cache, leading to an actual speed up that is considerably higher than the theoretical speedup.  These results demonstrate attractive practical value of AntMan on commodity hardware.

\section{Conclusion}
\edits{We develop AntMan, a technique that combines structured sparsity and low-rank decomposition to compress dense matrix-multiplications. We demonstrated its power to reduce the computation, size and execution time of RNN models by order(s) of magnitude, without losing accuracy. We hope its compression efficiency and effectiveness would help unblock and enable many great RNN-based models deployed in practice. 
AntMan also shines the lights on compressing
any matrix-multiplications based DL layers beyond RNNs, which is an interesting future work.
}

\clearpage

\bibliography{icml2019_conference}
\bibliographystyle{icml2019}

\end{document}